\documentclass[conference]{IEEEtran}
\usepackage{cite}
\usepackage{amsmath,amssymb,amsfonts}
\usepackage{algorithmic}
\usepackage{graphicx}
\usepackage{textcomp}
\usepackage[dvipsnames,table,xcdraw]{xcolor}
\usepackage{multicol}

\usepackage{amsmath}
\usepackage{amssymb}

\usepackage{multirow}
\usepackage{multicol}
\usepackage{makecell}
\usepackage{url}
\usepackage{booktabs}

\definecolor{orange}{rgb}{0.98823529, 0.85098039, 0.7372549}
\newcommand{\cc}{\cellcolor{orange}}

\def\BibTeX{{\rm B\kern-.05em{\sc i\kern-.025em b}\kern-.08em
    T\kern-.1667em\lower.7ex\hbox{E}\kern-.125emX}}
\begin{document}

\title{XiDepth: a Lightweight and Efficient Network for Self-supervised Monocular Depth Estimation}


\author{\IEEEauthorblockN{Elena Izzo}
\IEEEauthorblockA{\textit{Department of Mathematics} \\
\textit{University of Padova}\\
Padova, Italy \\
elena.izzo@phd.unipd.it}
\and
\IEEEauthorblockN{Riccardo Toniolo}
\IEEEauthorblockA{\textit{Department of Mathematics} \\
\textit{University of Padova}\\
Padova, Italy \\
riccardo.toniolo.2@studenti.unipd.it}
\and
\IEEEauthorblockN{Lamberto Ballan}
\IEEEauthorblockA{\textit{Department of Mathematics} \\
\textit{University of Padova}\\
Padova, Italy \\
lamberto.ballan@unipd.it}
}

\maketitle


\begin{abstract}
Self-supervised monocular depth estimation has emerged as an appealing solution to design lightweight and effective models for deployment on computationally constrained devices due to its reduced reliance on expensive depth sensors. By eliminating the need for ground-truth annotations and leveraging the simplicity of monocular camera setups, this approach facilitates cost-effective data collection and broad applicability across fields such as computer vision and robotics. A critical challenge is achieving resource-efficient neural networks without compromising the overall performance. State-of-the-art models generally adopt depth-wise convolutions and attention mechanisms; however, these functions often incur high energy costs and face compatibility issues in embedded environments.
To address this, we propose XiDepth, a lightweight architecture based on the XiNet operator block, designed to enhance feature extraction while maintaining low computational complexity and energy demand. On the KITTI dataset, XiDepth achieves state-of-the-art performance with only $0.8$M parameters. Tests on a Raspberry Pi 4 further confirm its suitability for real-world embedded applications, reducing FLOPs by $40$\% and energy consumption by $35$\% compared to leading methods.
\end{abstract}


\section{Introduction}
\label{sec:intro}

Depth Estimation is the task of measuring the distance of each pixel in an input image relative to the camera viewpoint. It is crucial for understanding the 3D world, and it is a key component in many applications in computer vision, autonomous driving, augmented reality, and robotics~\cite{qin2021monogrnet, zhang2021view, fan2022object, ziliotto2025tango}.
Traditionally, high-precision depth maps were obtained using LiDAR or stereo cameras. With the advancement in computer vision, attention shifted to deep-learning-based models using images from monocular cameras, which simplify equipment requirements and reduce cost with respect to their traditional counterparts. Initially, methods adopted a fully supervised learning strategy using LiDAR's ground truth as the supervised signal to regress depths~\cite{eigen2014depth, li2015depth}, but this requires expensive equipment to capture precise ground truth. To overcome this challenge, self-supervised methods propose to exploit geometric constraints between frames as the supervision signal, avoiding the need for large-scale annotated datasets~\cite{DBLP:journals/tip/WangBSS04, DBLP:conf/cvpr/CasserPMA19, DBLP:conf/iccv/JungPY21, DBLP:conf/cvpr/PoggiATM20, DBLP:conf/cvpr/YangSWC20, DBLP:conf/iccv/GodardAFB19}. 

In parallel, the recent interest in computationally constrained devices has underlined the need for lightweight, efficient, and scalable architectures. The main challenge is to obtain resource-efficient neural networks without compromising the overall performance. Generally, depthwise convolutions and the attention mechanism are widely used to optimize the performance-complexity trade-off~\cite{DBLP:conf/cvpr/ZhangNVK23, DBLP:conf/iccv/ZhouFSX21}, but they often incur high energy consumption and compatibility issues on embedded devices due to the use of non-standard operations~\cite{DBLP:conf/cvpr/SandlerHZZC18,tang2021bridge}. Since energy efficiency directly impacts battery life, thermal management, and system reliability, it is a critical factor for practical deployment. Recently, XiNet~\cite{DBLP:conf/iccv/AncilottoPF23} was introduced as a convolutional block optimized for efficiency in on-device execution and maximally exploiting the hardware's capabilities. However, its applicability is limited to image classification and object detection tasks, and its integration for dense prediction tasks is unexplored.

\begin{figure*}[th!]
    \centering
    \includegraphics[width=0.95\linewidth]{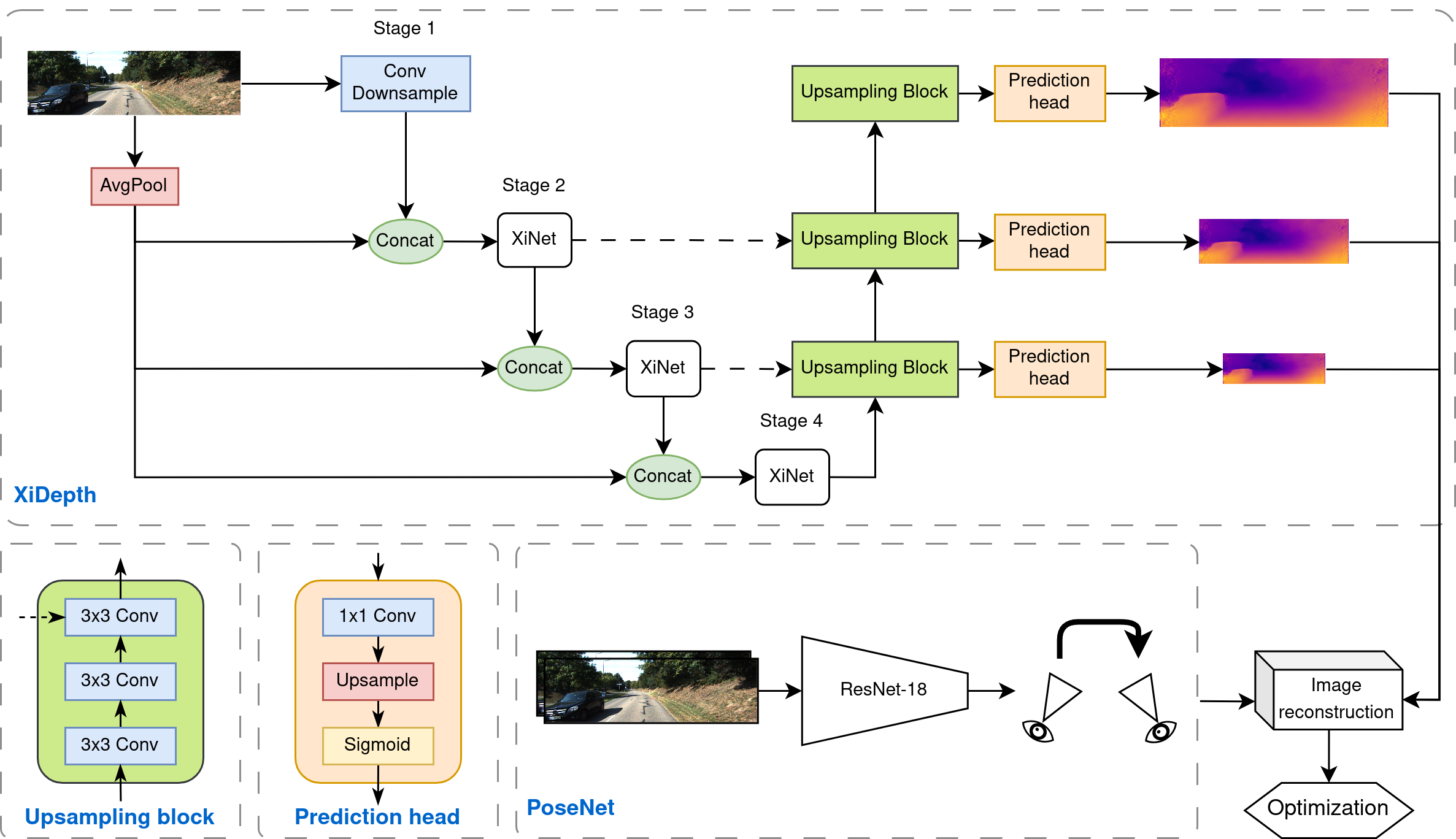}
    \caption{\textbf{Overview of the proposed framework.} The input image is processed via an encoder-decoder architecture to produce multiscale depth maps at full, $\frac{1}{2}$, and $\frac{1}{4}$ resolution. The encoder is based on the XiNet blocks to optimize energy consumption. The depth maps are used to reconstruct a time-closed frame of the input image, after estimating the relative pose with a PoseNet.}
    \label{fig:model_image}
\end{figure*}

In this paper, we propose XiDepth, a novel, lightweight, and energy-efficient self-supervised monocular depth estimation model based on the XiNet blocks. Our approach optimizes the complexity-performance trade-off by reducing the number of parameters, minimizing the computation overhead, and lowering the energy consumption. We validate our model on the KITTI dataset, where XiDepth achieves state-of-the-art performance with significantly fewer parameters. Moreover, tests on a Raspberry Pi 4 demonstrate superior energy efficiency, RAM and CPU usage, and inference time compared with existing solutions.

In summary, our contributions are: 
i) XiDepth, a lightweight architecture for self-supervised monocular depth estimation based on the XiNet block, tailored for energy- and resources-constrained devices;
ii) Competitive performance on the KITTI dataset with significantly fewer parameters than state-of-the-art methods;
iii) Evaluation on a Raspberry Pi 4, confirming advantages in energy efficiency, memory, and inference speed.


\section{Method}

Our XiDepth framework is depicted in Figure~\ref{fig:model_image}. It consists of a lightweight U-Net encoder-decoder network that estimates depth maps at multiple scales, while a dedicated pose-estimation network predicts the relative camera motion between adjacent frames. The estimated depth and pose are then combined to reconstruct neighboring views, generating the self-supervisory signal used during training. The encoder, described in~\ref{subsubsec:xidepth_encoder}, relies on the XiNet module for efficient on-device execution, avoiding depthwise convolutions to lower energy demand and non-standard operations to guarantee compatibility with embedded devices. The decoder estimates multi-scale inverse depth maps of the input image, then used by the Image Reconstruction module to reconstruct the target image and compute the loss (in ~\ref{subsubsec:xidepth_decoder}). To compensate for the reduced model capacity typically associated with lightweight networks, XiDepth leverages a self-supervised learning strategy based on view reconstruction. By exploiting geometric consistency across consecutive frames, the model learns meaningful depth representations without requiring ground-truth depth annotations, thereby reducing both annotation costs and deployment requirements (in~\ref{subsec:xidepth_sslstrategy}).

\subsection{XiDepth Architecture}
\subsubsection{XiDepth Encoder}
\label{subsubsec:xidepth_encoder}
The small encoder extracts multi-scale features across four stages. It takes an image  $x \in \mathbb{R}^{3 \times H \times W}$ as input, and makes the output prediction as follows:
\begin{equation}
    x_1 = \text{Conv}^{s1}_{3 \times 3}(\text{Conv}^{s1}_{3 \times 3}(\text{Conv}^{s2}_{3 \times 3}(x)))
\end{equation}
\begin{equation}
    x_i = \text{XiNet}_{L=2}(\text{concat}(x_{i-1},x^{AP}_{i-1})), \quad i=2,3,4
\end{equation}
\begin{equation}
    x^{AP}_{i} = \text{AP}_i(x), \quad \text{where} 
    \begin{cases}
        \text{AP}_n = \text{AP}(\text{AP}_{n-1}(x)) \\
        \text{AP}_1 = \text{AP}(x)
    \end{cases} 
\end{equation}
where Conv$^s_{k \times k}(\cdot)$ is a convolutional layer with stride $s$ and kernel size $k$; XiNet$_{L=2}$$(\cdot)$ is a XiNet module with 2 layers; concat$(\cdot)$ is the concatenation function between the output of the previous stage and  $x^{AP}_i$. The role of $x^{AP}_i$ is to reduce the loss of information due to the decreasing of the spatial resolution. It is obtained by applying $i$ consecutive average pooling layers AP$(\cdot)$ on the input image. 

The XiNet module makes the model more efficient from an energy point of view, avoiding dilated convolutions, and compatible with embedded devices due to the absence of non-standard operations.
The XiNet module takes a tensor $y \in \mathbb{R}^{C \times H \times W}$ as input, and makes the output as follows:
\begin{equation}
    z_0 = \text{Conv}^{s2}_{7 \times 7}(y)
\end{equation}
\begin{equation}
    z_{l} = \text{SiLU}(\text{BN}(\text{Conv}^{s1}_{3 \times 3}(\text{CompConv}^{s1}_{1 \times 1}(z_{l-1})) + s_l))
\end{equation}
\begin{equation}
    s_l = \text{CompConv}^{s1}_{1 \times 1}(\text{AAP}_l(z_1))
\end{equation}
where $z_l$ is the output of the $l$-th XiNet's convolutional block, CompConv$(\cdot)$ is a convolutional layer that compresses the output filters by a factor $\gamma$, BN$(\cdot)$ is the batch normalization layer~\cite{DBLP:conf/icml/IoffeS15}, AAP$_l(\cdot)$ is the adaptive average pooling layer, SiLU$(\cdot)$ and Sigmoid$(\cdot)$ are the activation functions.

\subsubsection{XiDepth Decoder}
\label{subsubsec:xidepth_decoder}
Once multi-scale features have been extracted by the encoder, a decoder estimates the inverse depth map at three scales: full, $\frac{1}{2}$, and $\frac{1}{4}$ resolution. To keep the decoder as compact as possible, we avoid a complicated up-sampling method~\cite{DBLP:conf/iccv/ZhouFSX21} and we privilege modules based on convolutions~\cite{DBLP:conf/cvpr/ZhangNVK23, DBLP:conf/iccv/GodardAFB19}. The decoder has three stages, each of them characterized by an upsampling block and a prediction head. The upsampling block increases the spatial dimension by applying bilinear upsampling on the concatenation of the input with one output feature from the encoder. Its output is passed to both the subsequent upsampling block and the prediction head for estimating the depth map.

\subsection{Self-supervised Learning Strategy}
\label{subsec:xidepth_sslstrategy}

Our goal is to train a model that predicts a depth map $D_t$ from a single color input $I_t$. To reach it, we adopt a self-supervised training strategy for monocular depth estimation. Following previous works~\cite{DBLP:conf/cvpr/ZhangNVK23, DBLP:conf/cvpr/ZhouBSL17}, the total loss $\mathcal{L}$ is computed as the mean of the sum of a photometric reprojection loss $\mathcal{L}_p$ and an edge-aware smoothness loss $\mathcal{L}_s$ obtained for each output scale $s$. Mathematically, it is as follows:
\begin{equation}
    \mathcal{L} = \frac{1}{3} \sum_{s \in \{1,\frac{1}{2}, \frac{1}{4}\} } (\mathcal{L}_p + \lambda\mathcal{L}_s)
\end{equation}
where $s$ represents the three depth scales output by the decoder, and $\lambda$ is a weighted parameter set to $1e^-3$ as in~\cite{DBLP:conf/iccv/GodardAFB19}.

\subsubsection{Photometric Reprojection Loss} The Photometric Reprojection Loss leverages the principle that, given two consecutive frames of the same scene, $I_t$ and $I_{t^{'}}$, and the relative pose between their viewpoints $P$, it is possible to estimate scene depth $D_t$ by reconstructing one view from the other. Consequently, starting from a target image $I_t$, a predicted depth map $D_t$, and the relative camera position $P$, one can synthesize the adjacent frame $I_{t^{'}}$.
This reconstruction enables self-supervised learning of depth without the need for ground-truth annotations. The key idea is to train the network to minimize the difference between the actual image $I_{t^{'}}$ and its reconstruction $\hat{I}_{t^{'}}$, using a photometric loss computed as:
\begin{equation}
    \hat{\mathcal{L}}_p(\hat{I}_{t^{'}}, I_{t^{'}}) = \alpha \dfrac{1 - SSIM(\hat{I}_{t^{'}}, I_{t^{'}})}{2} + (1-\alpha)\parallel\hat{I}_{t^{'}} - I_{t^{'}}\parallel_1
\end{equation}
where $SSIM(\cdot, \cdot)$ is the Structural Similarity Index~\cite{DBLP:journals/tip/WangBSS04}, $\alpha$ is set to 0.85~\cite{DBLP:conf/iccv/GodardAFB19}, and $\parallel \cdot \parallel_1$ is the $L1$ distance in pixel space.

The reconstructed image $\hat{I}_{t^{'}}$ is generated by a function $f(I_t, P, D_t, K)$ in terms of the input color image $I_t$, the estimated pose $P$, the depth $D_t$ predicted by the model, and the camera's intrinsics $K$.
The estimated relative pose $P$ is obtained by a PoseNet model composed of a pre-trained ResNet18~\cite{he2016deep} as the encoder and four convolutional layers as the decoder for computing the $6-$DoF relative pose between two consecutive images. This module is then removed at inference time.

Moreover, to deal with out-of-view pixels, occluded pixels, and moving pixels, we adopt the minimum photometric loss $\mathcal{L}^{min}_p$ strategy from~\cite{DBLP:conf/iccv/GodardAFB19}, formulated as:
\begin{equation}
    \mathcal{L}^{min}_p(\hat{I}_{t^{'}}, I_{t^{'}}) = \min_{I_{t^{'}} \in [I_{t+1}, I_{t-1}]} \hat{\mathcal{L}}_p(\hat{I}_{t^{'}}, I_{t^{'}})
\end{equation}
where $I_{t^{'}}$ can be the previous or the next frame with respect to the input image $I_t$. Additionaly,  we use a binary mask $\mu$ for removing pixels that violate camera motion assumptions:
\begin{equation}
    \mu = \big[ \mathcal{L}^{min}_p(\hat{I}_{t^{'}}, I_{t^{'}}) < \hat{\mathcal{L}}_p(\hat{I}_{t^{'}}, I_{t}) \big]
\end{equation}
Finally, the adjusted reprojection loss $\mathcal{L}_p$ is computed as:
\begin{equation}
    \mathcal{L}_p(\hat{I}_{t^{'}}, I_{t^{'}}) = \mu \odot \mathcal{L}^{min}_p(\hat{I}_{t^{'}}, I_{t^{'}})
\label{eq:final_repro_loss}
\end{equation}

\subsubsection{Edge-aware Smoothness Loss}
The edge-aware smoothness loss is used to smooth the estimated depth $D_t$ and it is computed as follows:
\begin{equation}
    \mathcal{L}_s = |\partial_x D^{*}_t|e^{-\partial_x I_t} + |\partial_y D^{*}_t|e^{-\partial_y I_t}
\end{equation}
where $D^{*}_t = D_t/\hat{D_t}$ is the mean-normalized inverse depth~\cite{DBLP:conf/cvpr/WangB0L18} to discourage values close to zero and increase training stability.


\begin{table*}[th!]
    \caption{\textbf{Comparison of XiDepth with some recent representative methods on the KITTI dataset using the Eigen split~\cite{DBLP:conf/iccv/EigenF15}}. The best is in \textbf{bold}, the second best is \underline{underlined}. All models are trained on the KITTI monocular videos with images resized to $640 \times 192$.}
    \centering
    \resizebox{\linewidth}{!}{
    \begin{tabular}{c|cccc|ccc|c}
        \multirow{2}{*}{\textbf{Method}} & \multicolumn{4}{c|}{\textbf{Depth Error ($\downarrow$)}} & \multicolumn{3}{c|}{\textbf{Depth Accuracy ($\uparrow$)}} & \textbf{Size ($\downarrow$)}\\
        \cline{2-9}
         & Abs Rel & Sq Rel & RMSE & RMSE log & $\delta < 1.25$ & $\delta < 1.25^2$ & $\delta < 1.25^3$ & Params. (M) \\ 

         \hline

         Monodepth2 & 0.132 & 1.044 & 5.142 & 0.210 & 0.845 & 0.948 & 0.977 & 14.3 \\
         R-MSFM3 & 0.128 & 0.965 & 5.019 & 0.207 & 0.853 & 0.951 & 0.977 & 3.5 \\
         R-MSFM6 & 0.126 & 0.944 & 4.981 & 0.204 & 0.857 & 0.952 & 0.978 &  3.8 \\
         Lite-Mono & \underline{0.121} & \textbf{0.876} & \underline{4.918} & \underline{0.199} & \underline{0.859} & \underline{0.953} & \textbf{0.980} & \underline{3.1} \\
         \cc XiDepth (Ours) & \cc \textbf{0.120} & \cc \underline{0.941} & \cc \textbf{4.876} & \cc \textbf{0.198} & \cc \textbf{0.870} & \cc \textbf{0.956} & \cc \textbf{0.980} & \cc \textbf{0.8} \\
         \bottomrule
    \end{tabular}
    }
    \label{tab:kitti_results}
\end{table*}

\begin{table*}
    \caption{\textbf{Complexity and efficiency comparison on a Raspberry Pi 4}. The best is in \textbf{bold}, the second best is \underline{underlined}. The last row shows the percentage difference between the best state-of-the-art value and our model.}
    \centering
    \resizebox{0.75\linewidth}{!}{
    \begin{tabular}{c|ccccc}
        \textbf{Method} & \textbf{\makecell{Inference \\ time (s)} ($\downarrow$)} & \textbf{\makecell{Energy \\ (mWh)} ($\downarrow$)} & \textbf{\%RAM ($\downarrow$)} & \textbf{\%CPU ($\downarrow$)} & \textbf{FLOPs (G) ($\downarrow$)} \\
         \hline

         Monodepth2 & 0.49 & \underline{80} & 13.42 & 97.01 & 8.0 \\
         R-MSFM3 & 79.41 & 124 & 14.00 & 96.71 & 16.5\\
         R-MSFM6 & 131.27 & 270 & \underline{13.30} & 96.83 & 31.1 \\
         Lite-Mono & \underline{0.48} & 83 & 14.53 & \underline{96.07} & \underline{5.0} \\
         \cc XiDepth (Ours) & \cc \textbf{0.31} & \cc \textbf{52} & \cc \textbf{13.00} & \cc \textbf{95.23} & \cc \textbf{3.0} \\
         \small \color{cyan}{$\Delta \%$} & \small \color{cyan}{$ -35 \%$} & \small \color{cyan}{$ -35 \%$} & \small \color{cyan}{$- 2 \%$} & \small \color{cyan}{$-  0.9 \%$} & \small \color{cyan}{$- 40  \%$} \\
         \bottomrule
    \end{tabular}}
    \label{tab:model_complexity}
\end{table*}

\begin{table*}
    \caption{\textbf{Generalization on the Make3D dataset.} All models are directly inferred on Make3D without fine-tuning.}
    \centering
    \resizebox{0.75\linewidth}{!}{
    \begin{tabular}{c|c|cccc}
        \textbf{Method} & \textbf{Training} & \textbf{Abs Rel} & \textbf{Sq Rel} & \textbf{RMSE} & \textbf{RMSE log} \\
        \hline

         DDVO & I+K & 0.387 & 4.720 & 8.090 & 0.204 \\
         Monodepth2 & I+K & 0.322 & 3.589 & 7.417 & 0.163 \\
         R-MSFM6 & I+K & 0.334 & 3.285 & 7.212 & 0.169 \\
         Lite-Mono & I+K & 0.305 & 3.060 & 6.981 & 0.158 \\
         \cc XiDepth (Ours) & \cc K & \cc 0.351 & \cc 3.925 & \cc 7.774 & \cc 0.175 \\
         \bottomrule
    \end{tabular}
    }
    \label{tab:make3d_results}
\end{table*}

\begin{figure*}[t]
    \centering
    \includegraphics[width=0.75\linewidth]{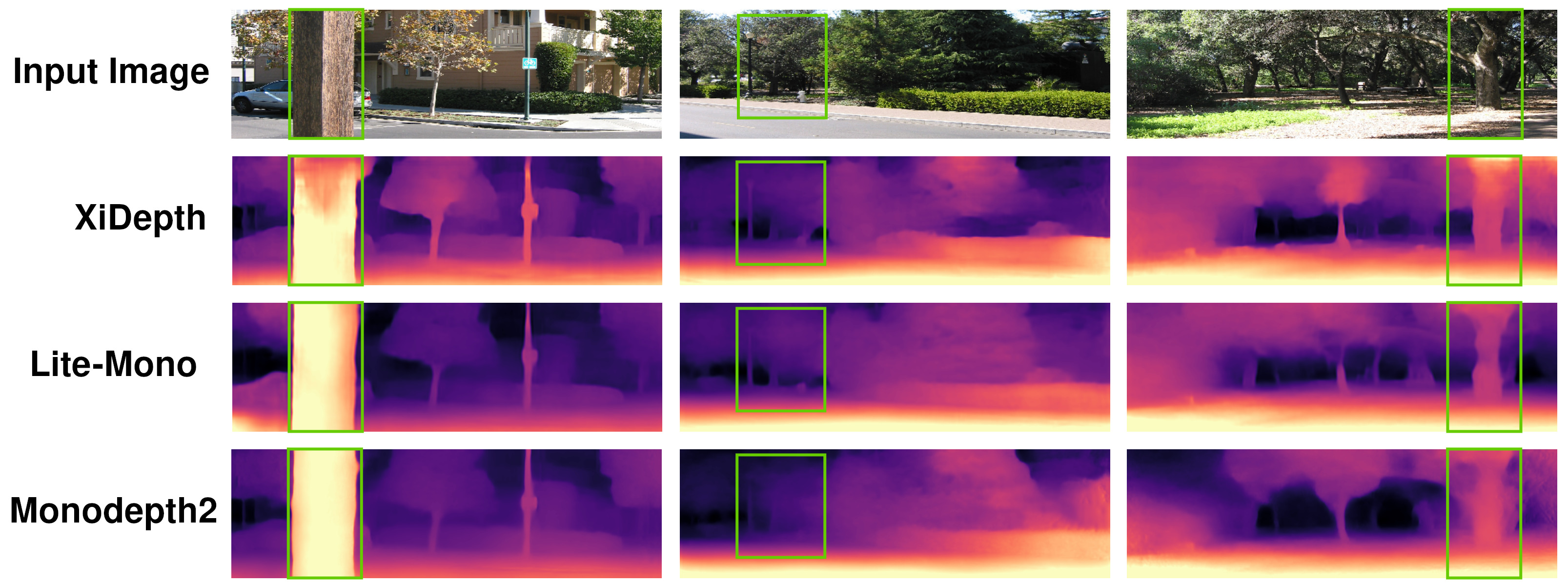}
    \caption{\textbf{Qualitative results on the Make3D dataset.} XiDepth is compared to Lite-Mono and Monodepth2.}
    \label{fig:make3d_visualization}
\end{figure*}

\section{Datasets and Implementation Details}

\subsection{KITTI Dataset}
The experiments were carried out using the KITTI dataset~\cite{DBLP:journals/ijrr/GeigerLSU13}. Its high-quality data collected from a driving perspective made it the de facto standard for single-frame depth evaluation. It contains about 42k stereo pairs composing 61 road scenes with multiple sensors, including, among the rest, camera, 3D Lidar, and GPU/IMU. Following the Eigen et al. split~\cite{DBLP:conf/iccv/EigenF15}, we use $39.810$ monocular triplets for training, $4.424$ for validation, and $697$ for testing. As in prior work~\cite{DBLP:conf/iccv/GodardAFB19}, we use the same camera intrinsic matrix $K$ for all images to compute the image reconstruction loss (Eq.~\ref{eq:final_repro_loss}) during training. It is obtained by averaging all the focal lengths of images across the dataset. 

\subsection{Make3D Dataset}
We further evaluate the generalization ability of our method on the Make3D dataset~\cite{DBLP:journals/pami/SaxenaSN09}. It contains 134 high-resolution RGB test images collected in outdoor scenes. Results on the Make3D dataset are obtained by directly inferring the model trained on KITTI. 

\subsection{Implementation Details}
We train the model for 300 epochs with a batch size of 12. We use AdamW as the optimizer, and a weight decay equal to $1e^{-2}$. The learning rate is set to $5e^{-4}$ with a cosine scheduler. Following~\cite{DBLP:conf/iccv/GodardAFB19, DBLP:conf/cvpr/ZhangNVK23, DBLP:conf/iccv/ZhouFSX21, DBLP:conf/aaai/LyuLWKLLCY21}, the data augmentation includes horizontal flips, brightness adjustment ($\pm 0.2$), saturation adjustment ($\pm 0.2$), contrast adjustment ($\pm 0.2$), and hue jitter ($\pm 0.1$), each applied with $50$\% probability and in a random order.

\section{Experimental Results}

We demonstrate the properties of the XiDepth model from both a performance and a complexity point of view. We first conduct studies on the challenging KITTI benchmark to quantify the generalization capabilities of our architecture. 
We compare our method with state-of-the-art works characterized by a self-supervised monocular training setting and models with less than $15$M parameters. We include both full CNN-based models and hybrid CNN-Transformer architectures in the analysis. Table~\ref{tab:kitti_results} reports the results on the KITTI dataset~\cite{DBLP:journals/ijrr/GeigerLSU13} across seven widely-used metrics~\cite{DBLP:conf/nips/EigenPF14}: Abs Rel, Sq Rel, RMSE, RMSE log, $\delta < 1.25$, $\delta < 1.25^2$, and $\delta < 1.25^3$. For all models, both training and evaluation are performed on the KITTI monocular videos with images resized to $640 \times 192$. The evaluation is performed on the Eigen split~\cite{DBLP:conf/iccv/EigenF15}, restricting the predicted depth range in $[0,80]$m, as is common practice. Our model obtains competitive results with the state-of-the-art ones, significantly reducing the number of parameters. In fact, XiDepth beats both Monodepth2~\cite{DBLP:conf/iccv/GodardAFB19} and R-MSFM~\cite{DBLP:conf/iccv/ZhouFSX21} in all metrics and outperforms or gets comparable results to Lite-Mono~\cite{DBLP:conf/cvpr/ZhangNVK23}.

Secondly, we test our model on a Raspberry Pi 4 to quantitatively compare the computational complexity and energy efficiency of our method against state-of-the-art approaches. Table~\ref{tab:model_complexity} reports the results in terms of inference time, energy consumption, RAM and CPU usage, and FLOPs (floating-point operations). During the measurement process, no active cooling was utilized on the device, and the only active process, aside from the operating system, was the Python script performing the measurements. The inference time is obtained as a mean of $100$ iterations on a single image with size $640 \times 192$. For computing both RAM and CPU usage, we mean values collected every 0.05 seconds. The energy consumption is obtained using an external current meter. 
Previous works estimate the efficiency of their models on indirect metrics, such as parameter count; however, these metrics are not the only indicator of the model's latency and its overall efficiency~\cite{DBLP:conf/iccv/AncilottoPF23}. As a result, despite both Lite-Mono~\cite{DBLP:conf/cvpr/ZhangNVK23} and R-MSFM~\cite{DBLP:conf/iccv/ZhouFSX21} being lightweight models with fewer than $4$M parameters, XiDepth significantly outperforms them in all metrics, gaining up to $35 \%$ in both energy and inference time and $40 \%$ in FLOPs. 

Finally, we evaluate the generalization capabilities of our model on the Make3D dataset~\cite{DBLP:journals/pami/SaxenaSN09}. Table~\ref{tab:make3d_results} reports the quantitative results, instead Fig.~\ref{fig:make3d_visualization} shows some qualitative examples. Unlike other approaches, which leverage large-scale ImageNet pretraining in addition to KITTI, XiDepth is trained exclusively on the KITTI dataset. Despite this, XiDepth achieves competitive performance in unseen outdoor environments, successfully capturing both local and global scene structures. We expect that incorporating large-scale pretraining in future work will further enhance cross-dataset generalization without compromising the lightweight and energy-efficient design of XiDepth.

\section{Conclusion}
In this work, we propose XiDepth, a novel, lightweight, and energy-efficient monocular depth estimation model. By integrating the XiNet blocks in a multiscale framework, our model effectively balances performance and computational efficiency.  On the KITTI dataset, our method outperforms previous works, significantly reducing the number of parameters and computational cost. Moreover, XiDepth largely beats its competitors in terms of energy consumption, RAM and CPU usage, and inference time on a Raspberry Pi 4, proving its outstanding energy-efficient properties.

\bibliographystyle{IEEEtran}
\bibliography{strings,refs}

\end{document}